\documentclass[twoside]{article}
\usepackage{ecj,palatino,epsfig,latexsym,natbib}
\usepackage{multirow}
\usepackage{amsmath}
\usepackage{amssymb}
\usepackage{rotating}
\usepackage{makecell}
\usepackage{url}
\usepackage{subcaption}
\usepackage{xcolor}

\renewcommand{\epsilon}{\varepsilon}

\usepackage{mathtools}

\newcommand{\R}{\mathbb{R}}

\usepackage{algorithm}
\usepackage{multirow}
\usepackage{hyperref}
\usepackage{algorithmic}

\begin{document}

\ecjHeader{x}{x}{xxx-xxx}{202X}{Illuminating the Diversity-Fitness Trade-Off in Black-Box Optimization}{M. L. Santoni, E. Raponi, A. Neumann, F. Neumann, M. Preuss, C. Doerr }
\title{\bf Illuminating the Diversity-Fitness Trade-Off in Black-Box Optimization}  

\author{\name{\bf Maria Laura Santoni} \hfill \addr{maria-laura.santoni@lip6.fr}\\ 
        \addr{Sorbonne Universit\'e, CNRS, LIP6, Paris, France}
\AND
       \name{\bf Elena Raponi} \hfill \addr{e.raponi@liacs.leidenuniv.nl}\\
        \addr{Leiden Institute of Advanced Computer Science, Universiteit Leiden, 2333 CC Leiden, The Netherlands}
\AND
       \name{\bf Aneta Neumann} \hfill \addr{aneta.neumann@adelaide.edu.au}\\
        \addr{Optimisation and Logistics,
School of Computer and Mathematical Sciences,
The University of Adelaide, Adelaide SA, 5005, Australia}
\AND
       \name{\bf Frank Neumann} \hfill \addr{frank.neumann@adelaide.edu.au}\\
        \addr{Optimisation and Logistics,
School of Computer and Mathematical Sciences,
The University of Adelaide, Adelaide SA, 5005, Australia}
\AND
       \name{\bf Mike Preuss} \hfill \addr{m.preuss@liacs.leidenuniv.nl}\\
        \addr{Leiden Institute of Advanced Computer Science, Universiteit Leiden, 2333 CC Leiden, The Netherlands}
\AND
       \name{\bf Carola Doerr} \hfill \addr{carola.doerr@lip6.fr}\\
        \addr{Sorbonne Universit\'e, CNRS, LIP6, Paris, France}
}

\maketitle

\begin{abstract}
In real-world applications, users often favor structurally diverse design choices over one high-quality solution. It is hence important to consider more solutions that decision makers can compare and further explore based on additional criteria. Alongside the existing approaches of evolutionary diversity optimization, quality diversity, and multimodal optimization, this paper presents a fresh perspective on this challenge by considering the problem of identifying a fixed number of solutions with a pairwise distance above a specified threshold while maximizing their average quality.

We obtain first insight into these objectives by performing a subset selection on the search trajectories of different well-established search heuristics, whether they have been specifically designed with diversity in mind or not. We emphasize that the main goal of our work is not to present a new algorithm but to understand the capability of off-the-shelf algorithms to quantify the trade-off between the minimum pairwise distance within batches of solutions and their average quality. We also analyze how this trade-off depends on the properties of the underlying optimization problem.

A possibly surprising outcome of our empirical study is the observation that naive uniform random sampling establishes a very strong baseline for our problem, hardly ever outperformed by the search trajectories of the considered heuristics. We interpret these results as a motivation to develop algorithms tailored to produce diverse solutions of high average quality. 
\end{abstract}

\section{Introduction}
\label{sec:intro}

In various real-world applications, relying solely on a single optimal parameter configuration can lead to unexpectedly poor performance. For instance, in engineering design, the solution obtained from an optimizer might not be accurately translated into a mechanical prototype due to manufacturing or cost constraints that have not been taken into account in the modeling of the problem. Additionally, function evaluations heavily depend on numerical simulations, which may be affected by numerical inconsistencies or simplifications. As a consequence, the actual quality of a real prototype can fall below its expectation. 
It is then crucial to provide alternative solutions to designers, enabling them to make choices based on diverse criteria~\citep{raponi_kriging-assisted_2019,dommaraju_identifying_2019,yousaf_similarity-driven_2023,bujny_learning_2023}.
Unfortunately, certain critical criteria for selecting an optimal solution cannot be directly translated into a diversity metric and, therefore, cannot be embedded within the optimization procedure. Examples include ease of manufacturing, which depends on the available machinery and the characteristics of pre-manufactured parts or materials that are readily available from suppliers, aesthetic considerations related to the elegance of the design, and robustness, defined as stable performance under slight variations in configuration due to manufacturing errors. Furthermore, presenting a diverse set of solutions in terms of parameter configurations can lead to aesthetically varied layouts, among which stakeholders may select a preferred option based on subjective impressions derived from visual inspection. Thus, generating a diverse batch of solutions enables engineers to use their domain expertise post hoc, and to incorporate additional subjective or qualitative criteria into the decision-making process.

We can look at this scenario as a slightly reshaped optimization problem. Rather than looking for a single, good or even optimal, solution, we are interested in strategies capable of identifying high-quality solutions in the search space that are characterized by the best possible average fitness while also satisfying a minimum distance criterion. As soon as we derive a solution set instead of a single solution, the choice of criteria describing the performance of the set becomes ambiguous. This is complicated by the fact that usually we have no information about the number of optima or their distribution in search and objective space.

Similar problems have already been addressed in several subfields of optimization. 
\textit{Evolutionary diversity optimization (EDO),} for instance, works with a fixed quality threshold and evolves sets of solutions that all meet the given threshold and have maximal diversity according to a given diversity measure for the population. For EDO, different indicators for measuring the diversity in high-quality solution sets have been investigated, e.g., star discrepancy \citep{DBLP:conf/gecco/NeumannGDN018}, and well-established indicators from the field of evolutionary multi-objective optimization \citep{DBLP:conf/gecco/NeumannG0019}.
\textit{Quality Diversity (QD)} is a popular search paradigm in robotics and games \citep{DBLP:journals/firai/PughSS16,DBLP:journals/tec/CullyD18,DBLP:conf/cig/GravinaKLTY19,DBLP:conf/cig/AlvarezDFT19}, which aims to illuminate the space of solution behaviors by exploring diverse niches in the feature space through partitioning, and simultaneously maximizing quality within each niche~\citep{mouret2015illuminating,vassiliades2017using,DBLP:conf/cig/AlvarezDFT19}.
\textit{Multimodal optimization (MMO)}, also referred to as \textit{niching}, is an approach that seeks to identify concurrently multiple discernible solutions for a multimodal objective function. This is achieved by exploring various modes or peaks in the search space, with each mode representing a distinct, typically locally optimal, solution~\citep{Preuss2021}.

We look at the trade-off between diversity of the solutions and their fitness from yet another perspective. Our key guiding question is as follows. \textit{Which solutions are available with the best possible average fitness but at a pre-defined pairwise distance? And how does this trade-off evolve when evaluated on concrete optimization problems of different nature?} Note that this is conceptually different from the perspective adopted in the aforementioned fields. 
The rationale behind EDO relies on defining a fixed quality threshold that the returned solutions must meet while maximizing a diversity measure, whereas we are not only interested in finding solutions that are \textit{good enough} but rather \textit{as good as possible} under pre-defined distance requirements.

To ensure diversity, QD operates in the behavioral space rather than the search or objective space (without going into details, this ``behavorial space'' can be thought of as some kind of intermediate, typically low-dimensional space specifically chosen to evaluate the diversity of the solutions). QD aims to cover as much of the behavioral space as possible, providing a diverse set of high-performing solutions that exhibit distinct behaviors. In contrast to this, our diversity requirement concern the pairwise distance in the \textit{search space}, also referred to as \textit{design space} in engineering domains. 
Finally, in MMO, the number of solutions returned depends on the problem multimodality, and there are no \textit{pre}defined criteria for diversity.

\textbf{Our contribution: }
In the absence of a dedicated algorithm tailored to our quest of returning solution batches balancing diversity and quality under a pre-defined distance requirement, this work explores the feasibility of using established black-box optimization techniques to generate portfolios $X$ of points from which these high-quality batches of some predefined size $k$ can be extracted. Here, we define a \textit{batch} as a subset of $k$ candidate points chosen to balance diversity and quality within the search space, while a \textit{portfolio} refers to a larger set of points---eventually an optimization history (i.e., a search \textit{trajectory})---serving as the pool from which these batches are extracted.
We benchmark various algorithms specifically designed for MMO tasks, as they were originally intended for objectives that are close to ours, as well as non-adaptive sampling strategies (random and Sobol' sampling) and the well-known covariance matrix adaptation evolution strategy (CMA-ES) \citep{hansen_reducing_2003}. 

To evaluate the potential of trajectories of points generated by existing optimization approaches for this novel subset selection problem, we define baselines on top of the 24 noiseless BBOB test functions from the COCO environment~\citep{hansen2021coco} for dimension 2, 5, and 10.
The BBOB problem set is specifically designed to include various representative landscapes commonly encountered in numerical black-box optimization. Therefore, it is an especially fitting choice for our study, where we analyze the effects of different algorithm point archives and batch sizes for our subset selection problem. Our intention is not to propose a new algorithm to best tackle the diversity-fitness trade-off challenge. Instead, our aim is to offer a new perspective and undertake a preliminary investigation into the interaction of various factors, such as batch and portfolio size, enforced distances, and average solution quality, across different problem landscapes. As a long-term goal, we aspire to develop algorithms that provide users with control over this trade-off.

\textbf{Structure of the paper:} The problem definition and the basic approaches to address it are summarized in Section~\ref{sec:prob}. Results obtained from massive random sampling, allowing us to investigate the impact of the evaluated points to choose from, as well as that of the sought batch size, are discussed in Section~\ref{sec:first}. In the same section, we also provide a visualization of the obtained solutions for different enforced distances (Figure~\ref{isoall0}). Section~\ref{sec:Comparison} presents the results for the portfolios obtained from state-of-the-art multimodal algorithms and their comparison with those obtained with quasi-random sampling, uniform sampling, and CMA-ES. We conclude the paper in Section~\ref{sec:conclusion}.

\textbf{Reproducibility and supplementary material:} Our code for reproducing the experiments is available on GitHub~\citep{code}. The data used for plotting and the supplementary materials are available in the Zenodo repository \url{https://zenodo.org/doi/10.5281/zenodo.10992322}.

\section{Selecting Diverse Subsets using Algorithm Trajectories}\label{sec:prob}
\subsection{Problem Statement}

\textbf{The general problem.} 
Considering minimization as objective, we state our problem as follows. Given a (possibly unknown) function $f:S\subseteq  \R^D\rightarrow \R$, a minimum distance requirement~$d_{\min}>0$, and a batch size~$k\ge 1$, find a collection $X^*=\{x^1,\ldots, x^k\} \subseteq S$ such that the pairwise (Euclidean) distance between any two points $x^i$ and $x^j$ with $i \neq j$ satisfies $d(x^i,x^j) \geq d_{\min}$ and such that the average objective value $\sum_{i=1}^k f(x^i)/k$ is as small as possible. 

One may exchange the role of the constraints on the batch size and on the minimum enforced pairwise distance with the objective, and ask for maximizing the minimum distance that can be enforced while respecting an average quality constraint and the batch size requirement. Similarly, one may ask for maximizing the number of points that have pairwise distance at least $d_{\min}$ and whose average objective value is below a given threshold. While we see room for exploring all three variants of the problem, for the sake of brevity, we mostly focus on the one explicitly formulated above, but our data also sheds light on the other cases.

We acknowledge that both $d_{\min}$ and $k$ are problem parameters that can be challenging to define a priori. In our analysis, we explore how the trade-off between diversity and quality evolves for different values of these parameters.

\textbf{Constructing portfolios via subset selection.} Since we are not aware of any algorithm that directly addresses the above-stated problem, we will base all our analyses on the following approach. We first obtain an \textit{initial portfolio} $X$ of $T$ evaluated points. This portfolio serves as a pool of potential solutions from which we aim to extract an optimal subset $X^*\subset X$ of $k$ points that meets a minimum distance requirement and has high average quality. Formulating the problem in this way allows us to avoid searching the entire, potentially vast, search space directly, which would be impractical. We define the subset selection problem as follows:
\begin{equation}
\label{eq:problem_statement}
\begin{aligned}
\arg\min_{X^* \subset X} \quad & \sum_{x\in X^*} f(x)/k \\
\text{s.t.} \quad & |X^*| = k \\
& x^i, x^j \in X^* \implies d(x^i,x^j) \geq d_{\text{min}}, \quad \forall i \neq j. \\
\end{aligned}
\end{equation}

In our context, the portfolio $X$ contains the points generated by (adaptive or not) black-box optimization algorithms. 
Assuming that we could optimally solve the subset selection problem~\eqref{eq:problem_statement}, our task is to identify initial portfolios $X$ that contain high-quality subsets. 
However, subset selection, that is finding $X^*$, is by itself a notoriously difficult problem. 
We can therefore not hope to solve it optimally, especially without constraining portfolio sizes $T$ and dimensions $D$ to small values. Therefore, we also investigate heuristic strategies to approximate this solution. The specific approaches we use to tackle this problem are detailed in the following section.

\subsection{Solving the Subset Selection Problem}
\label{sec:Challenge}

\textbf{An exact approach using Gurobi.} Our first approach to obtain the subsets is an exact subset selection approach implemented in the \verb|Gurobi|~\citep{gurobi} library. Gurobi is a mathematical programming library designed for solving various optimization problems, including linear programming, quadratic programming, and mixed-integer programming. Its efficiency and accuracy make it a valuable tool for obtaining precise optimal solutions. However, this exact strategy is extremely time-consuming and memory-intensive, even for a single minimal distance calculation, making it infeasible to purely rely on it. 

\textbf{A greedy heuristic.} To obtain results for a broader range of minimal distances, we employ a straightforward greedy heuristic to approximate the batch $X^*$. This algorithm iteratively chooses batches of solutions from a large portfolio, ensuring that, at each iteration, the pairwise distance between points in the batch increases.

The underlying idea of this approach is as follows: considering a black-box function $f$ to be minimized, a portfolio $X^0$ (where the superscript $0$ denotes the initial iteration) of $T$ points, and a batch size $k$, we iteratively select points in the batch $B$ (approximating $X^*$) that minimize the average function value. We first sort the points in $X^0$ by increasing fitness values. Abusing notation, we call the resulting sequence $\bar{X^0}$. 
We start the heuristic by computing the minimum pairwise distance $\bar{d}_{\text{min}} = \min\{d(x,y) \mid x,y \in B^0, x \neq y\}$ between the points in the first batch $B^{0}$, which consists of the $k$ best points in $X^0$. For the two points defining this minimal distance, we identify the optimal set achievable by removing one of them and substituting it with another point $\bar{x}$ from the initial sample. For each of the two points, a candidate substitute is selected, i.e., the first point in the sequence $\bar{X^0}\setminus B^0 $ whose minimal pairwise distance to any of the points in the current batch is at least as large as before.
The set is then updated by including the point whose inclusion leads to the lowest average fitness in the new batch $\bar{B^0}$. This process iterates for a given maximal number $M$ of iterations. In our experiments, $M$ is always set to $1\,000$. However, for the settings $T = 1\,000\,000$ and $k = 10$ in $D = 5$ and $10$, the runs were stopped after four days, to keep the computational effort reasonable.

\textbf{Computational resources.}
We run our experiments on a Linux cluster~\citep{lrz} with a 64-bit x86 CPU architecture. The cluster is equipped with Intel Xeon CPU E5-2690 v3 processors running at 2.60 GHz. It consists of 812 nodes, with 28 cores and 64 GB of memory per node.

\begin{figure*}[h!]
    \center
     \includegraphics[width=\textwidth]{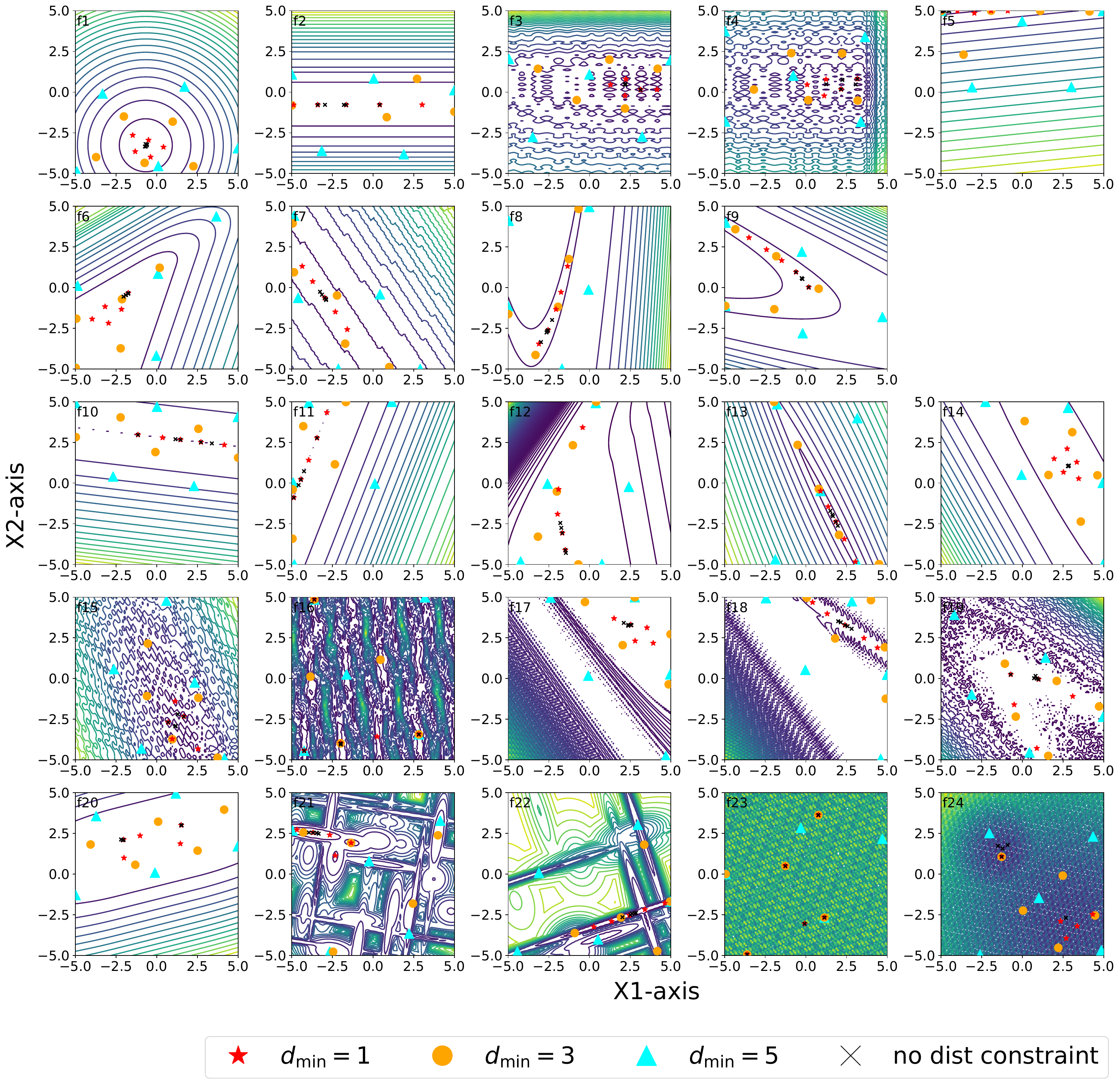}
      \caption{$D = 2$. Isocontours of the 24 BBOB functions with $k = 5$ Gurobi solutions, obtained from an initial set of $T = 10\,000$ points sampled uniformly at random. Showing results for $d_{\text{min}}$ = 1, 3, 5 and the batch of solutions with no distance constraint. Lighter colors correspond to worse solutions, the optima are hence in regions surrounded by dark lines.}
    \label{isoall0}
\end{figure*}

\section{First Results for the BBOB Function Suites}\label{sec:first}

\textbf{The BBOB functions.} 
As motivated in the introduction, we examine the evolution of the diversity-fitness trade-off for the 24 noiseless BBOB functions of the COCO environment~\citep{hansen2021coco}. In the experiments presented in this work, we consider the problems with instance ID = 0. The functions are defined on the search space $[-5,5]^D$. For convenience of integrating the optimization algorithms considered in the subsequent sections, we access the BBOB functions through IOHprofiler~\citep{doerr2018iohprofiler}.  

The BBOB functions are divided into five groups: separable functions (f1-f5), functions with low or moderate conditioning (f6-f9),
functions with high conditioning and unimodal structure (f10-f14), multimodal functions with appropriate (f15-f19), and weak global structure (f20-f24). The plots for the 24 BBOB functions presented in the following sections are organized according to their respective groups: functions from group one are placed in the first row, those from group two in the second row, and so on. This arrangement facilitates the comparison of performance across the different function groups.

\textbf{Performance criteria.} An advantage of the BBOB functions is that we know the value of the optimal solution. Algorithm performance is hence often measured in terms of \textit{target precision,} defined as the absolute difference between the best solution quality obtained by an algorithm and that of the global minimum. In our study we average the target precision of the $k$ solutions obtained for a given distance requirement $d_{\min}$. We refer to this average target precision as the \textit{loss}.

\subsection{Visualizations of Obtained Batches in \texorpdfstring{$2D$}{2D}} 
\begin{figure}[t]
    \center
    \includegraphics[width=.8\columnwidth]{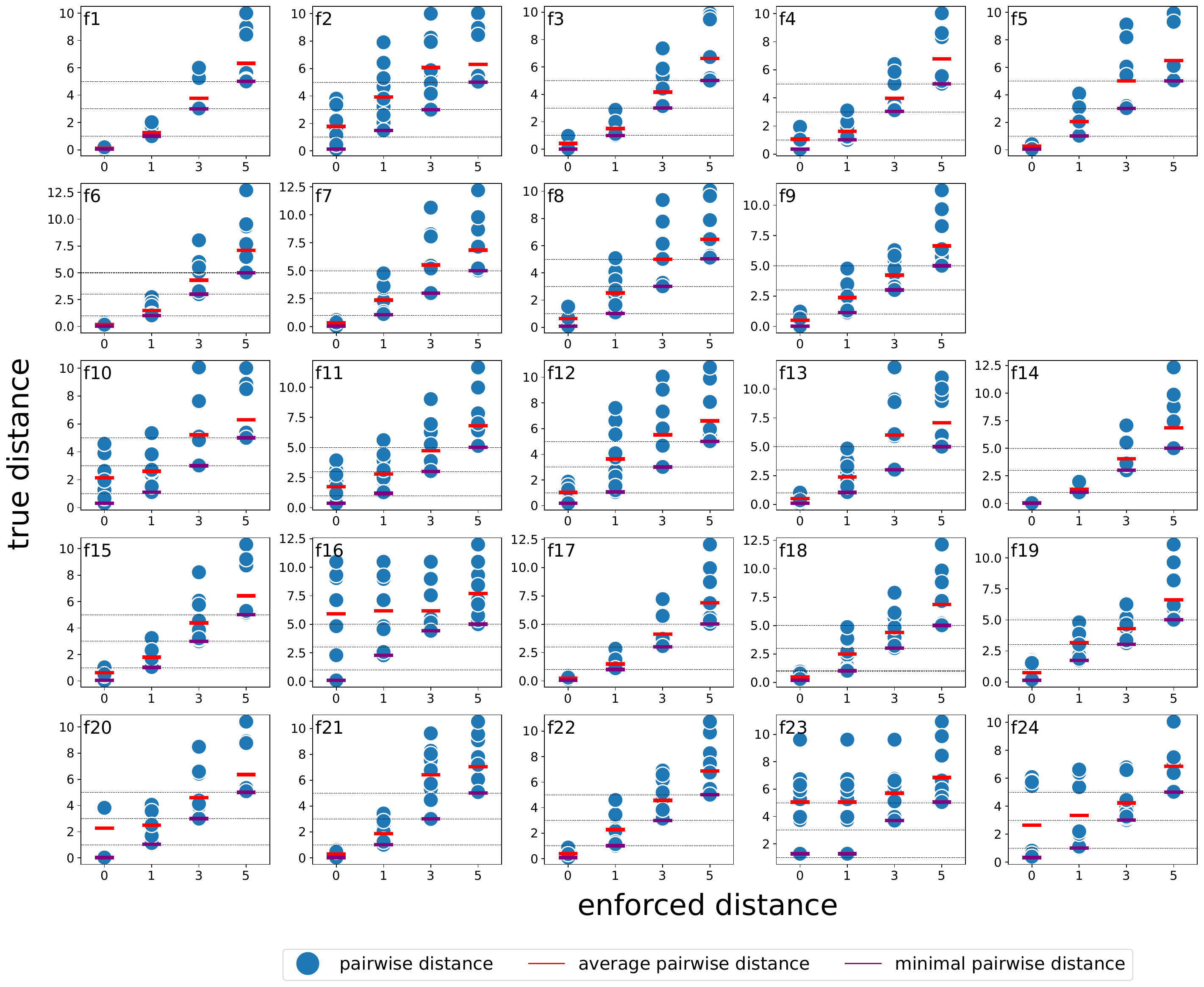}
      \caption{$D = 2$. Pairwise distances between the $k = 5$ points selected by Gurobi for the 24 BBOB functions, initiated from random samples of size $T = 10\,000$.}
    \label{distancesgurobi}
\end{figure}

Our first step is to illustrate the $2$-dimensional point sets that can be obtained via the exact Gurobi approach, applied to an initial set of $T = 10\,000$ points sampled uniformly at random in the search space. We use a batch size of $k = 5$ and we consider three different distance thresholds, $d_{\text{min}}$ = \{1, 3, 5\}. We add the batch of the $k$ best points (without distance constraint). The results are visualized in Figure~\ref{isoall0}. We immediately see (confer, for example, function f23) that the true distance between points can be much larger than the requested threshold. To investigate these effects, we plot in Figure~\ref{distancesgurobi} the distributions of the pairwise distances of the selected batches. Varying behaviors of the batches can be observed, contingent on the different landscape characteristics.

The unimodality of functions f1 to f14 (except for f3 and f4) induces a monotonic increase in the average pairwise distance between points in the optimal batches found with Gurobi as the enforced distance increases, as illustrated in Figure~\ref{distancesgurobi}. This behavior is anticipated, as larger enforced distances imply moving points farther away from the single basin of attraction, generally in the direction of the gradient. 
In f10 and f11, high ill-conditioning allows optimal points to spread along the largest direction of the contour line without sacrificing quality, even without distance constraints. Consequently, Figure~\ref{distancesgurobi} shows only minimal differences in the true average distance for no and small enforced distances.

Despite being multimodal, functions f15, f17, f18, and f19 exhibit a global unimodal structure with added noise. Consequently, the optimal batch behaves similarly as it does in unimodal functions, for increasing enforced distances (visible in both Figure~\ref{isoall0} and Figure~\ref{distancesgurobi}).
A distinct behavior is instead evident for multimodal functions with a repetitive landscape. In such cases, multiple basins of attraction exist, each nearly equivalent in fitness, and distributed across the entire search space (refer to Figure~\ref{isoall0}). This implies that, regardless of the minimum enforced distance, the optimal subset is already composed of points with a naturally high true pairwise distance. This results in a stable marker distribution across enforced distances, as depicted in Figure~\ref{distancesgurobi}. Examples illustrating this behavior are functions f16 and f23.

Finally, different observations about the trade-off between diversity and fitness can be drawn for multi-modal functions with a weak global structure. These functions are characterized by a local structure around the basins of attraction that differs from the structure presented globally in the search space. This distinction is reflected in the distribution of optimal points in the batches, which typically undergo a sharp change in true pairwise distance at a certain enforced distance value. This is evident for f20, f21, and f24 in Figure~\ref{distancesgurobi}, where a noticeable change in the distance distribution occurs between enforced distances of 1 and 3.
In such cases, the optimal batch tends to propagate within the global attraction basin as much as the enforced distance allows. It may either move in the gradient direction (f20) or jump to other attraction basins (f21 and f24) afterward. Function f24 is particularly interesting because its two basins of attraction have almost the same degree of optimality. Therefore, smaller clusters in the optimal subsets are created for no or small enforced distances, highlighted by the empty space between markers in Figure~\ref{distancesgurobi}.
All in all, it is crucial to distinguish between landscapes where diversity can naturally manifest without compromising the quality of optimal solutions, as they are already naturally distant from each other, and scenarios where constraints must be applied to ensure that optimal solutions maintain sufficient diversity from each other.

\begin{figure*}[h!]
    \centering
    \begin{subfigure}[b]{\textwidth}
        \centering
        \includegraphics[width=\textwidth]{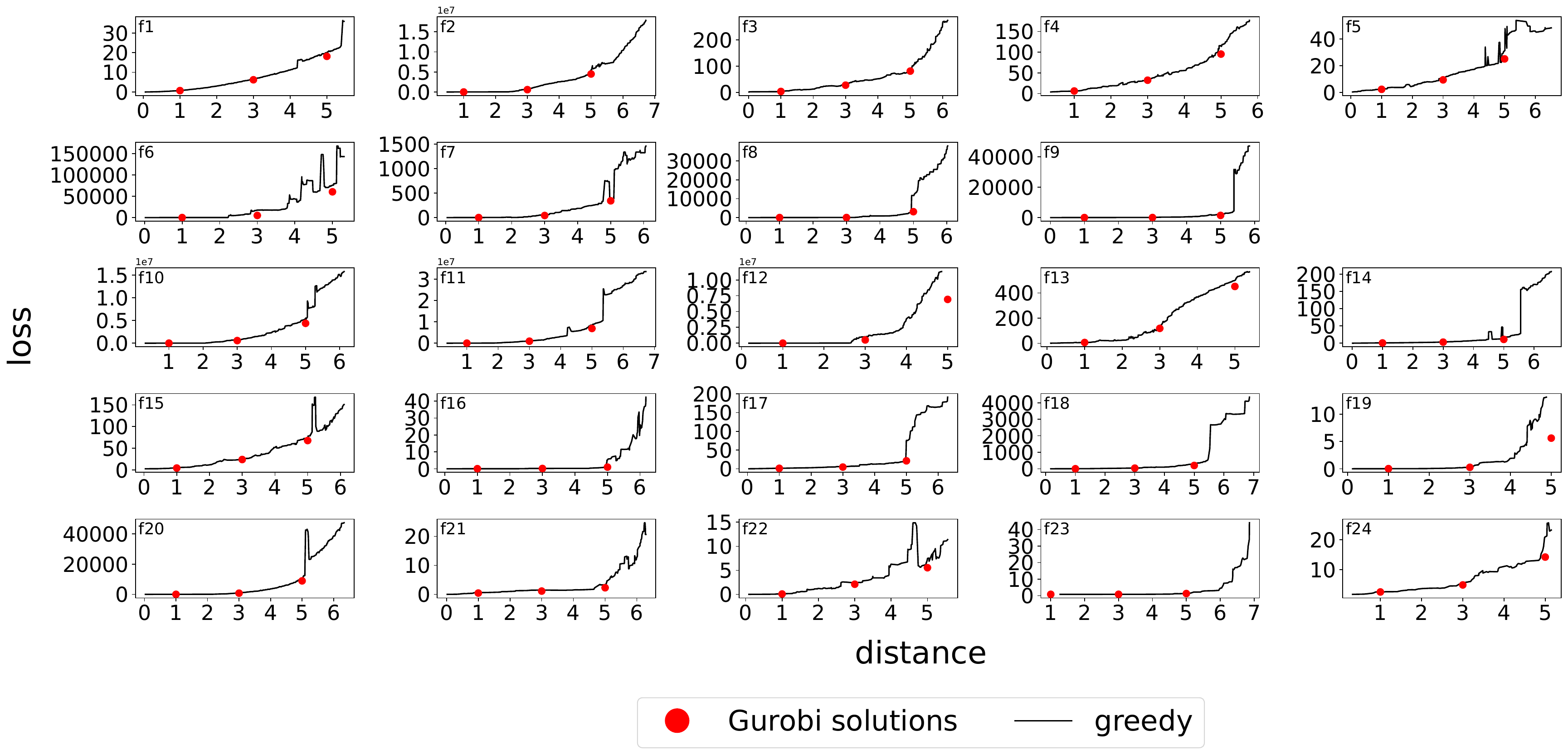}
        \caption{$D=2$, $d_{\text{min}}$ = 1, 3, 5.}
        \label{Gurobid2}
    \end{subfigure}
    \vspace{0.5cm} 

    \begin{subfigure}[b]{\textwidth}
        \centering
        \includegraphics[width=\textwidth]{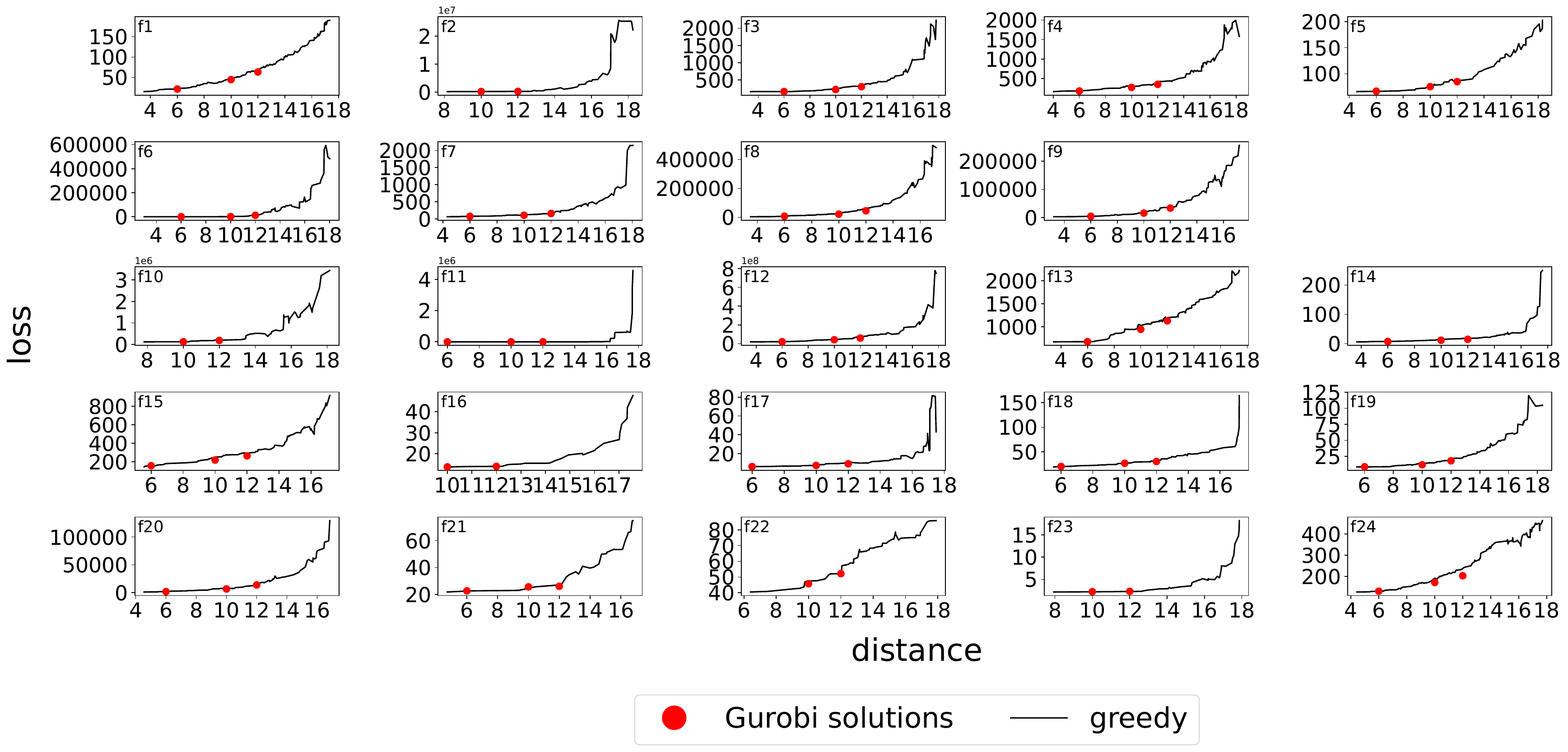}
        \caption{$D=10$, $d_{\text{min}}$ = 6, 10, 12.} 
        \label{Gurobid10}
    \end{subfigure}

    \caption{One run of the greedy approach, applied to an initial set of  $T = 10\,000$ points sampled uniformly at random and evaluated on the 24 BBOB functions in $D=2$ (a) and $D=10$ (b). Results are for batches of $k=5$ solutions. The red dots show the results obtained with Gurobi, applied to the same initial sets and different distances, $d_{\text{min}}$ = 1, 3, 5 (a) and $d_{\text{min}}$ = 6, 10, 12 (b) (distance 6 is included where feasible).}
    \label{Gurobi}
\end{figure*}

\subsection{Trade-off between Loss and Minimal Distance}
In Figure~\ref{Gurobi}, we present the trade-off between loss and $\bar{d}_{\text{min}}$ resulting from the execution of the greedy algorithm. Note that we do not manually correct the non-monotonic behavior of the greedy approach, to recall its heuristic nature. Such a post-processing would be easy to do, though, and can be easily obtained for the visual inspection of our results, since we always show the full curves, and one could replace the plotted curves by their best-so-far results (the ``lower envelope''). 

The red dots in Figure~\ref{Gurobi} represent the loss values for batches computed by Gurobi under $d_{\text{min}} = 1, 3, \text{ and } 5$ for the 2-dimensional case, and $d_{\text{min}} = 6, 10,$ and $12$ for the 10-dimensional case. For the 10-dimensional case, functions where the initial $\bar{d}_{\text{min}}$ in $B^{0}$ was already greater than 6 do not show the respective red dots for Gurobi. Please note that for $D = 10$, we checked up to $d_{\text{min}} = 12$ because going beyond this value would lead to violations of memory and time constraints of our cluster. The analogous figure for the 5-dimensional case is provided in the supplementary material. 

Our primary focus is obtaining the optimal solutions represented by the red dots computed by Gurobi. The figure empirically demonstrates that the much less computationally expensive greedy approach provides a strong approximation of the Gurobi solutions for both $D = 2$ and $D = 10$, with only marginal deterioration as $\bar{d}_{\text{min}}$ increases, as observed in functions f6, f12, f13, and f19 for $\bar{d}_{\text{min}}$ = 5 in the 2-dimensional case, and in function f24 for $\bar{d}_{\text{min}}$ = 12 in the 10-dimensional case. 
This underscores the efficiency of the greedy algorithm as a viable and more resource-efficient alternative for approximating optimal solutions for our specific task. Therefore, we proceed with our analyses by utilizing the greedy approach to conduct evaluations for various settings, i.e., different combinations of function, instance, dimension, sampling algorithm, initial portfolio size, and batch size.

\subsubsection{Impact of the Initial Set Size \texorpdfstring{$T$}{T}}
\label{sec:K5}

\begin{figure}[t]
    \center
    \includegraphics[width=\textwidth]{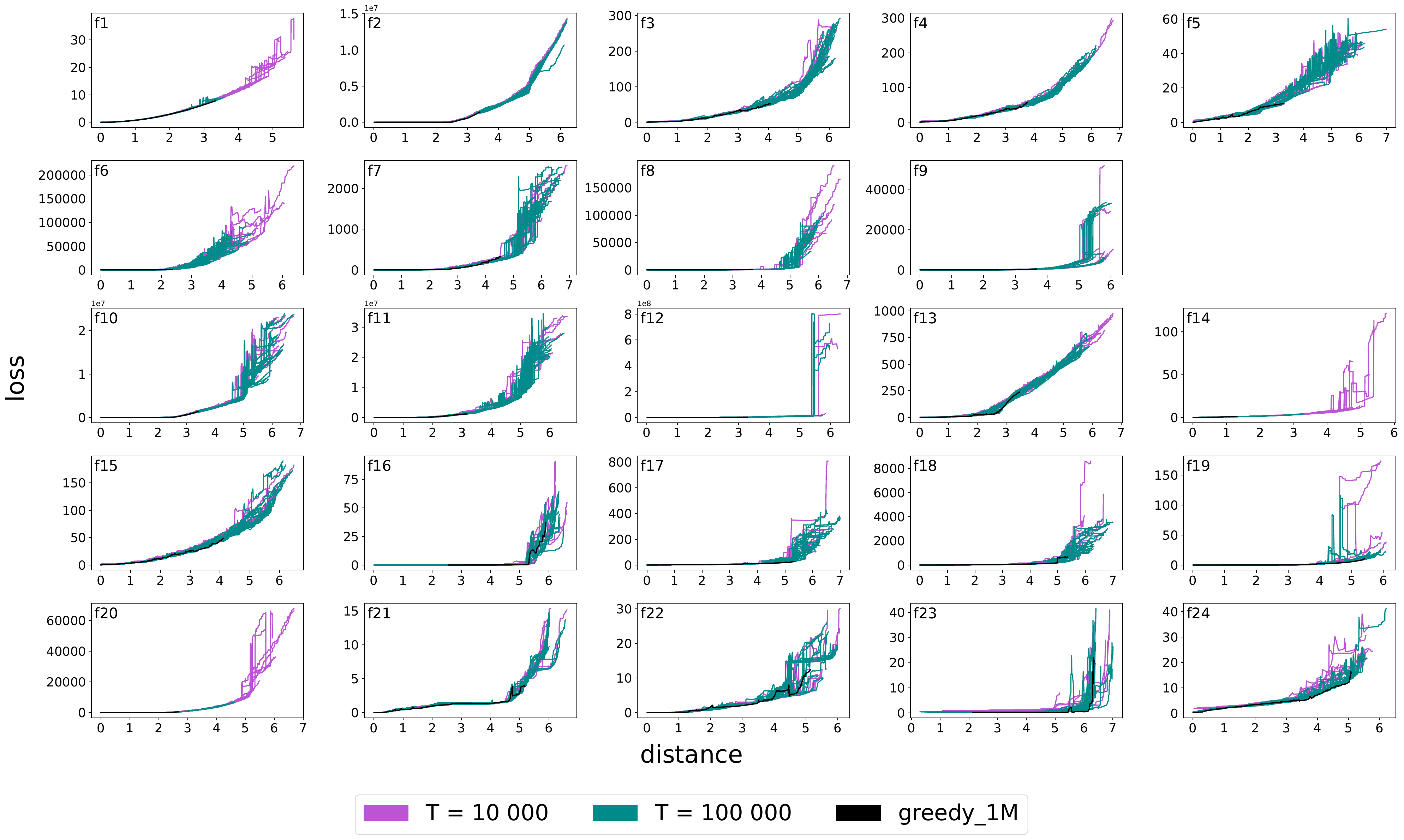}
      \caption{$D = 2.$ Impact of the size $T$ of the initial set on the distance-quality trade-off. Results are for the 24 BBOB functions, using a batch size of $k=5$, and using initial sets that are sampled uniformly at random. 10 and 20 independent repetitions for $T = 10\,000$ and $T = 100\,000$, respectively, and one run for an initial set of $T = 1\,000\,000$ points.
}
    \label{k5}
\end{figure}

In Figure~\ref{k5}, we investigate the impact of the portfolio size $T$ in the greedy approach. For $D=2$, the performances obtained for an initial sample size of 10\,000, 100\,000, and 1\,000\,000 points are comparable, suggesting that a relatively small portfolio is sufficient in low-dimensional spaces to achieve a satisfactory coverage of the solution space. Using excessively large portfolios in such cases leads to negligible improvements in solution quality while increasing computational costs. 

As dimensionality increases, however, the portfolio size becomes more critical. In higher-dimensional spaces (e.g., $D=10$), as the volume of the search space grows exponentially with the dimension, larger portfolios provide better coverage and improve both approximation quality and solution diversity. This observation is further explored in Section~\ref{sec:D10}, where Figures~\ref{d10k5log} and~\ref{d10k51000log} show these improvement. 

However, based on these findings and to manage computational overhead, for the subsequent comparison in Section~\ref{sec:Comparison} we fix the portfolio size to $T=10\,000$ for all dimensions ($D=2,5,$ and $10$) even though it ideally should scale with dimensionality.

\subsubsection{Impact of the Batch Size \texorpdfstring{$k$}{k}}
\begin{figure*}[t]
    \center
    \includegraphics[width=\textwidth]{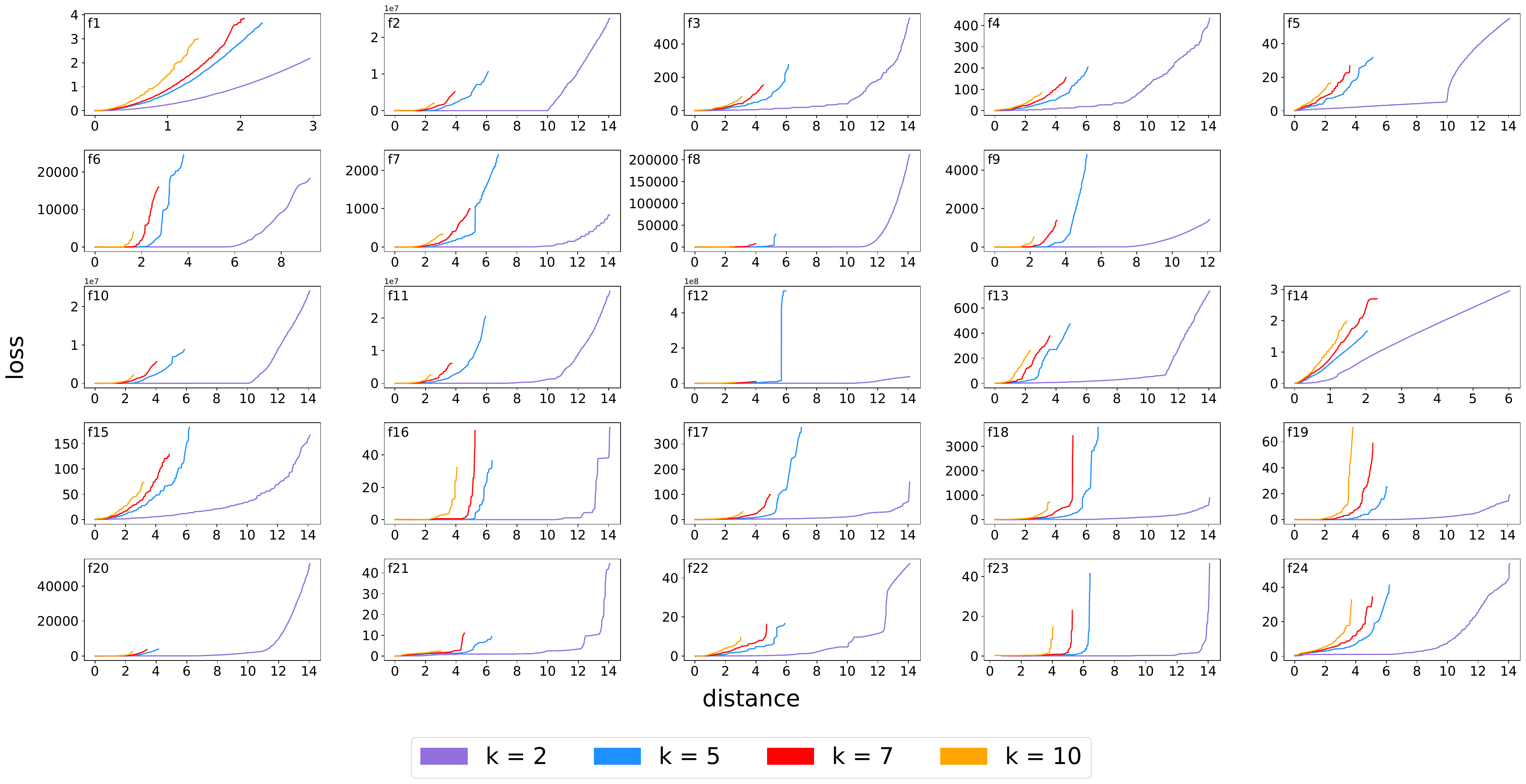}
      \caption{Evolution of the loss, plotted against minimum distance in the optimal batch for the 24 BBOB functions for $D=2$. Curves are generated by extracting the lower envelope from 10 independent runs for $k = 2, 7$ and 20 runs for $k = 5, 10$ of the greedy approach, initiated from a random sample of size $T = 100\,000$.
      }
    \label{T100k}
\end{figure*}

We use a batch size of $k = 5$ for our analyses. However, we acknowledge this parameter's selection is application-dependent. 
The desired number of returned solutions directly influences the loss of the batch, with an increase in this number leading to a deterioration in loss, as depicted in Figure~\ref{T100k}. The figure shows results for dimension 2, where batches are selected from an initial portfolio of $T = 100\,000$ points. The marked difference between $k = 2$ and the other values highlights the increased flexibility allowed by the distance constraint when the batch size is small relative to the search space dimensionality. While this pattern is evident in our example with $D = 2$, further investigation would be needed to confirm whether it extends to higher dimensions. The distinct jump in performance around distance equal to $10$ is attributed to the bounds of our space, $S=[-5,5]^D$, which forces the points to align in the diagonal direction at $D=2$. Exceptions to this pattern are observed in functions f19, and f21--f24, which, for the considered instance, either present basins of attraction more closely aligned with the diagonal or exhibit a highly repetitive landscape.

\section{Diversity-fitness Trade-off: Algorithm Benchmarking}
\label{sec:Comparison}

To the best of our knowledge, the existing literature lacks algorithms specifically designed to address our diversity-fitness trade-off objective. To assess the potential of existing algorithms in providing good initial sets, we employ our greedy algorithm on portfolios of points generated from either non-adaptive sampling strategies (sometimes referred to as \emph{one-step} sampling) or as histories resulting from runs of well-established heuristics for single-objective optimization or multimodal optimization (MMO) algorithms. 

\subsection{Algorithms}
\label{sec:algorithms}

We benchmark on the 24 noiseless BBOB functions diverse algorithms designed for MMO, also known as \textit{niching}, given the proximity between the task they were designed for and our ultimate goal. In fact, MMO algorithms aim to find several distinct solutions of a complex objective function simultaneously by means of exploring different modes or peaks in the search space~\citep{Preuss2021}. We include in our analysis the winners from the `Niching Methods for Multimodal Optimisation' competition series at the Genetic and Evolutionary Computation Conference from 2016 to 2019, by taking their most recent implementations. We benchmark the three top-ranked MMO algorithms~\citep{Preuss2021}: the \textit{Hill-Valley Evolutionary Algorithm (HillVallEA)}~\citep{maree2019benchmarking}, the \textit{Covariance Matrix Self-Adaptation with Repelling Subpopulations (RS-CMSA)}~\citep{article}, and the \textit{Weighted Gradient and Distance-Based Clustering Method (WGraD)}~\citep{9002742}. 

We compare the ability of these three algorithms in providing good portfolios for our greedy approach with the one of other baselines: an initial set sampled either uniformly at random or with a Sobol' sequence, and the well-known single-objective optimization heuristic Covariance matrix adaptation evolution strategy (CMA-ES)~\citep{hansen_reducing_2003}. This diverse set of initializations aims to explore how various techniques perform in finding solutions that not only optimize the objective function but also differ significantly from one another. The aim is to assess the quality of evaluations produced during the execution of MMO algorithms in terms of their suitability for the objective of our study.

We use default settings for the algorithm hyperparameters. As we did throughout the paper, we evaluate algorithm performance in terms of loss, i.e., average target precision of the obtained batches. The implementation for the random sample uses the \verb|random.uniform| method from the Python module \verb|numpy|~\citep{harris2020array}. We generate the Sobol' sequences via the method \verb|i4_sobol_generate| from the Python package \verb|sobol_seq|~\citep{sobol}.
The implementation of CMA-ES is the one available from the GitHub repository \verb|pycma|~\citep{hansen_reducing_2003}. The default settings do not include restarts.
The code for HillvallEA is taken from the GitHub repository \verb|HillVallEA|~\citep{maree2019benchmarking}.
For RS-CMSA, we also use the implementation distributed by the author~\citep{article}.
The code for WGraD is taken from the GitHub repository \verb|WGraD|~\citep{9002742}.

\subsection{Results}
\label{sec:results}

In this section, we explore $D=2$ and $D=10$ to assess the algorithms' effectiveness in finding diverse, high-quality batches of solutions as the dimensionality of the problem increases. Plots for $D = 5$ are provided in the supplementary material.

For each sampling method, we executed 10 independent runs limiting the maximum number of evaluations to $T = 10\,000$ and $T = 1\,000$. We also conducted experiments for $T = 100$; detailed results can be found in the supplementary material. Given the deterministic nature, we executed a single run of the Sobol' sequence sampling method. The resulting point portfolios are then used as a starting point for our greedy approach. 
The largest portfolio size ($T = 10\,000$) is selected following the analysis in Section~\ref{sec:K5} and considering that a greater number of points requires more CPU time for the greedy algorithm to generate batches for higher minimum pairwise distances.
We show results for $k=5$ for all dimensions. The results for $k=10$ can be found in the supplementary material.
In all plots, we show the trade-off obtained by running the greedy algorithm on a portfolio of $T = 1\,000\,000$ randomly generated points, represented by a solid black line. 

\subsubsection{Results for Dimension \texorpdfstring{$D = 2$}{D=2}}
\label{sec:D2}

\begin{figure}[t]
    \center
        \includegraphics[width=\textwidth]{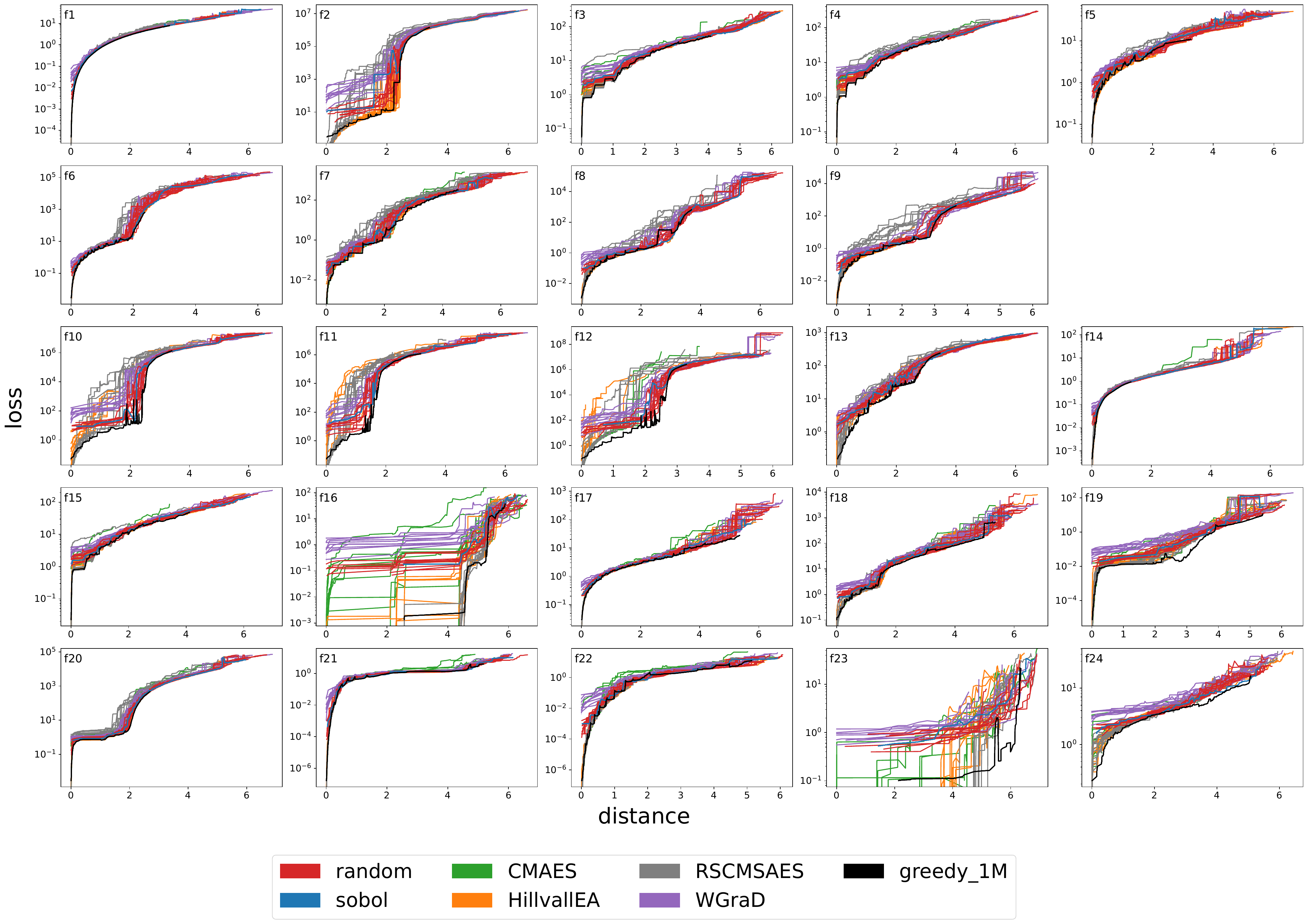}
      \caption{$D=2$. Trade-off between loss and minimum distance for an optimal batch of $k = 5$ points. 10 independent runs of the greedy approach starting from different point portfolios of size $T = 10\,000$.}
    \label{d2k5log}
\end{figure}

\begin{figure}[h]
    \center
    \includegraphics[width=\textwidth]{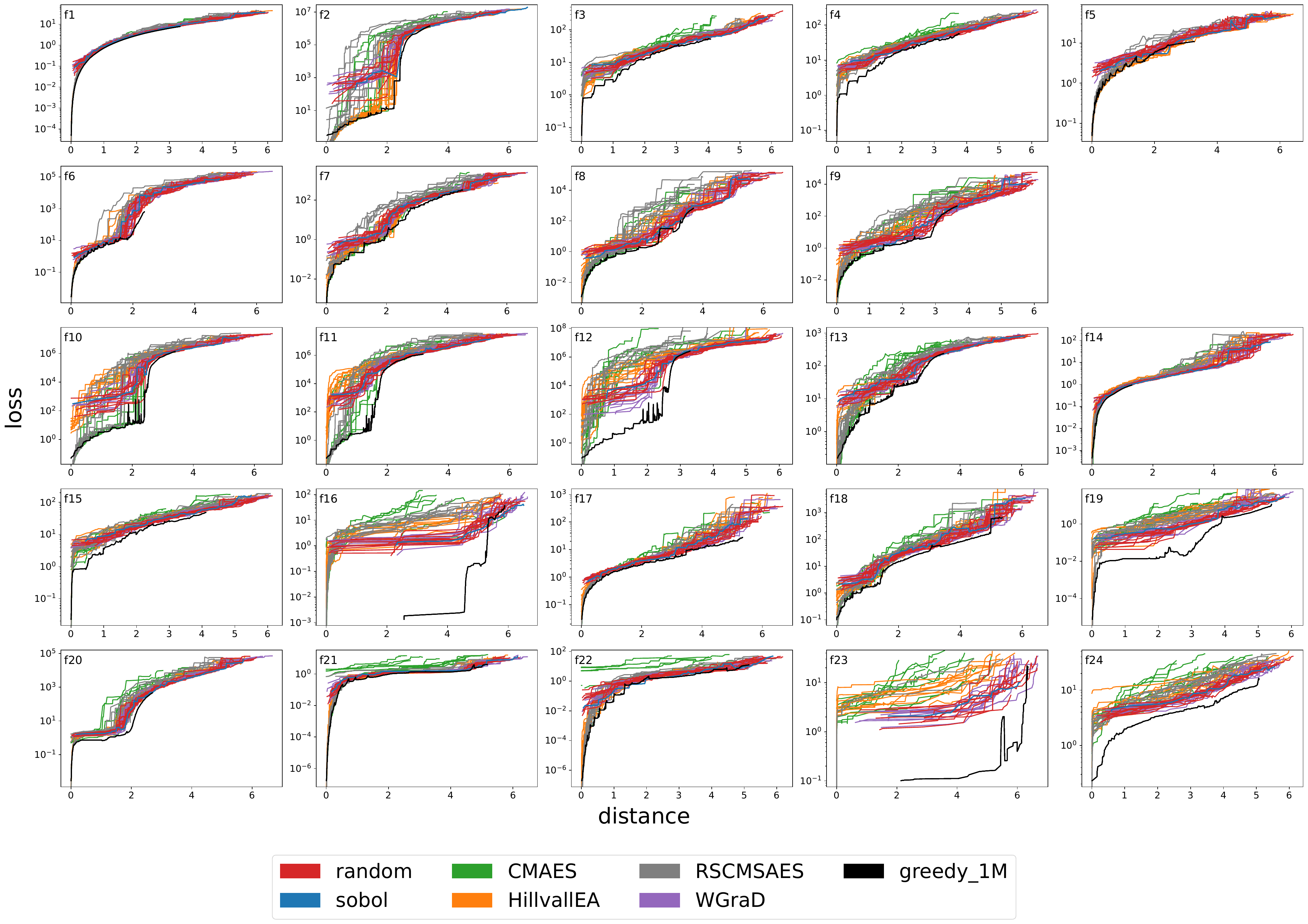}
      \caption{
      $D=2$. Same as Figure \ref{d2k5log}, but for initial portfolio size $T = 1\,000$.}
    \label{d2k51000log}
\end{figure}

In Figure~\ref{d2k5log} we can see that an initial portfolio of $10\,000$ points in dimension $D=2$ provides a good coverage of the search space regardless of the strategy that is used to generate them. This is reflected by the similar trade-offs between loss and minimum distance in the optimal batches across methods. 
The greedy algorithm, starting from a random sample, consistently performs surprisingly well, exhibiting only slight deviations from its performance when initiated with a portfolio of one million points sampled uniformly at random, which we consider as our baseline.

We observe improved performance from the MMO algorithms, specifically HillvallEA and RS-CMSA, as well as CMA-ES, for enforced distances typically smaller than~$1$. If the distance is sufficiently small, allowing the solutions in the optimal batch to be concentrated in one area --- either the global attraction basin or a small number of niches of equal quality --- MMO algorithms and CMA-ES offer a more extensive selection to the greedy algorithm. This is thanks to their tendency to locate more points in promising areas of the search space in their optimization histories. As the enforced distance increases, random sampling becomes more effective, since it distributes solutions uniformly across the domain.

Always due to their convergence capabilities, the same three optimization algorithms perform well on f16 and f23, which are characterized by very repetitive landscapes and many attraction basins of similar quality. In this case, HillVallEA and RS-CMSA are superior to CMA-ES as they are better-suited for multimodal optimization. 

This changes when we reduce $T$ to $1\,000$. In Figure~\ref{d2k51000log}, on f16 and f23, the algorithm that best approaches the performance of the greedy baseline is random sampling, although it falls short of achieving an optimal trade-off. 

With the smaller budget allocated to the sampling strategies, more pronounced differences emerge between the various algorithms.
Non-adaptive sampling strategies exhibit greater robustness, while optimization strategies show a variance depending on the run. For functions f2, f10-f13, f16, f20-f24, CMA-ES tends to localize the search (with no restarts by default). This effect is less common for niching algorithms, as they are typically not misled by local basins of attraction.

For smaller portfolios, the superiority of MMO algorithms and CMA-ES over random sampling becomes more evident for small enforced distances, typically up to around distance equal to $2$. This phenomenon is particularly clear for most of the functions with high conditioning and unimodal structure (f10-f13), where the histories of the optimizers locate a substantial portion of good-quality points along the largest direction of the contour line. This allows for keeping a good loss when increasing the enforced distance. 

In Figures~\ref{d2k5log} and Figure~\ref{d2k51000log}, some functions (f2, f6, f8--f13, f18, f20) present an evident elbow shape. This is either due to severe ill-conditioning or large optimal basins, which allow to spread points in such a way that they keep a good quality while being pushed far apart from each other. It is evident that these landscapes are well-suited for returning batches of good-quality and diverse solutions to the user, rather than focusing solely on a single optimal solution.

\subsubsection{Results for Dimension \texorpdfstring{$D = 10$}{D=10}}
\label{sec:D10}

\begin{figure}[t]
    \center
    \includegraphics[width=\textwidth]{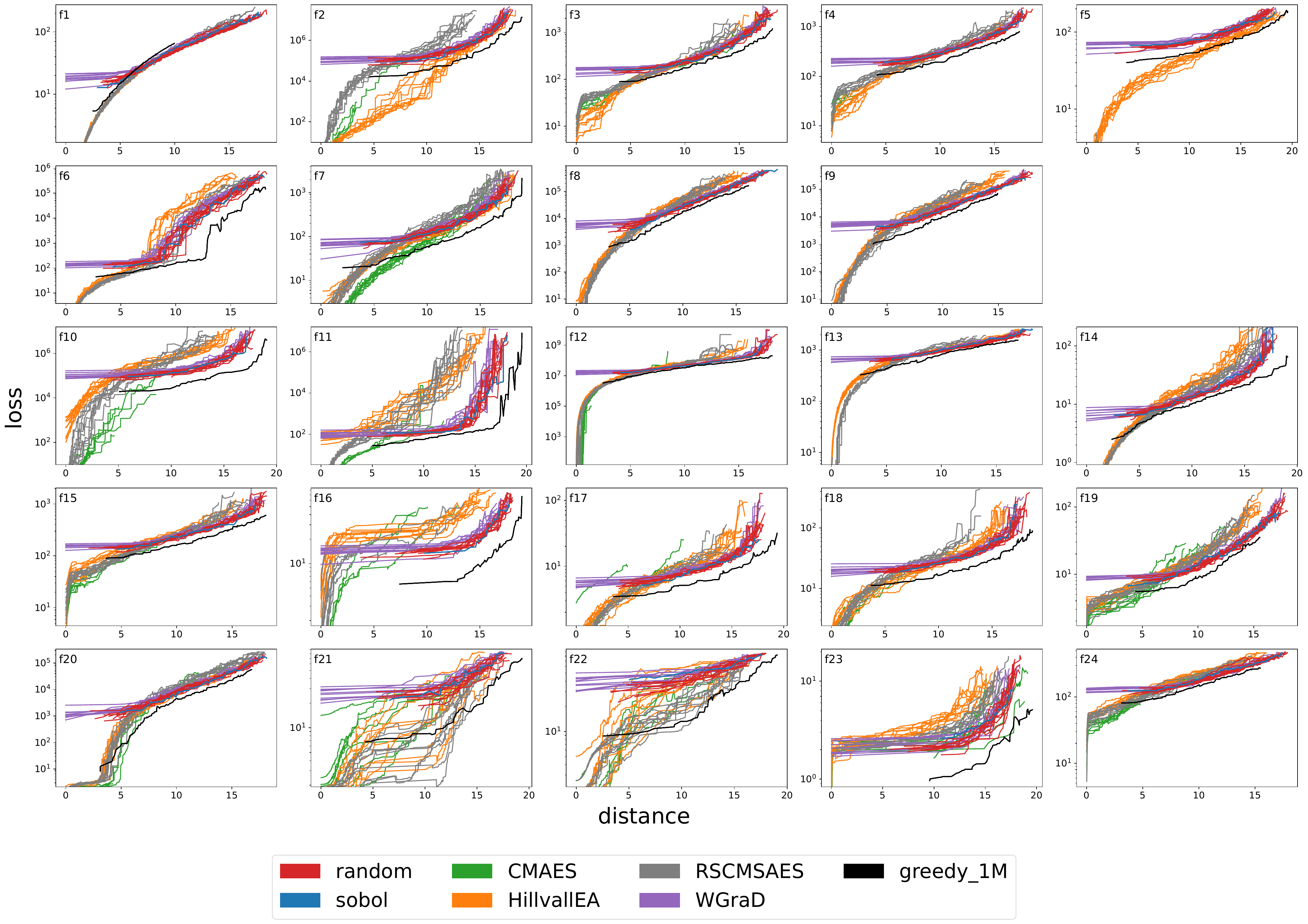}
      \caption{$D=10$. Trade-off between loss and minimum distance for an optimal batch of $k = 5$ points. 10 independent runs of the greedy approach starting from different point portfolios of size $T = 10\,000$.}
    \label{d10k5log}
\end{figure}

\begin{figure}[h]
    \center
    \includegraphics[width=\textwidth]{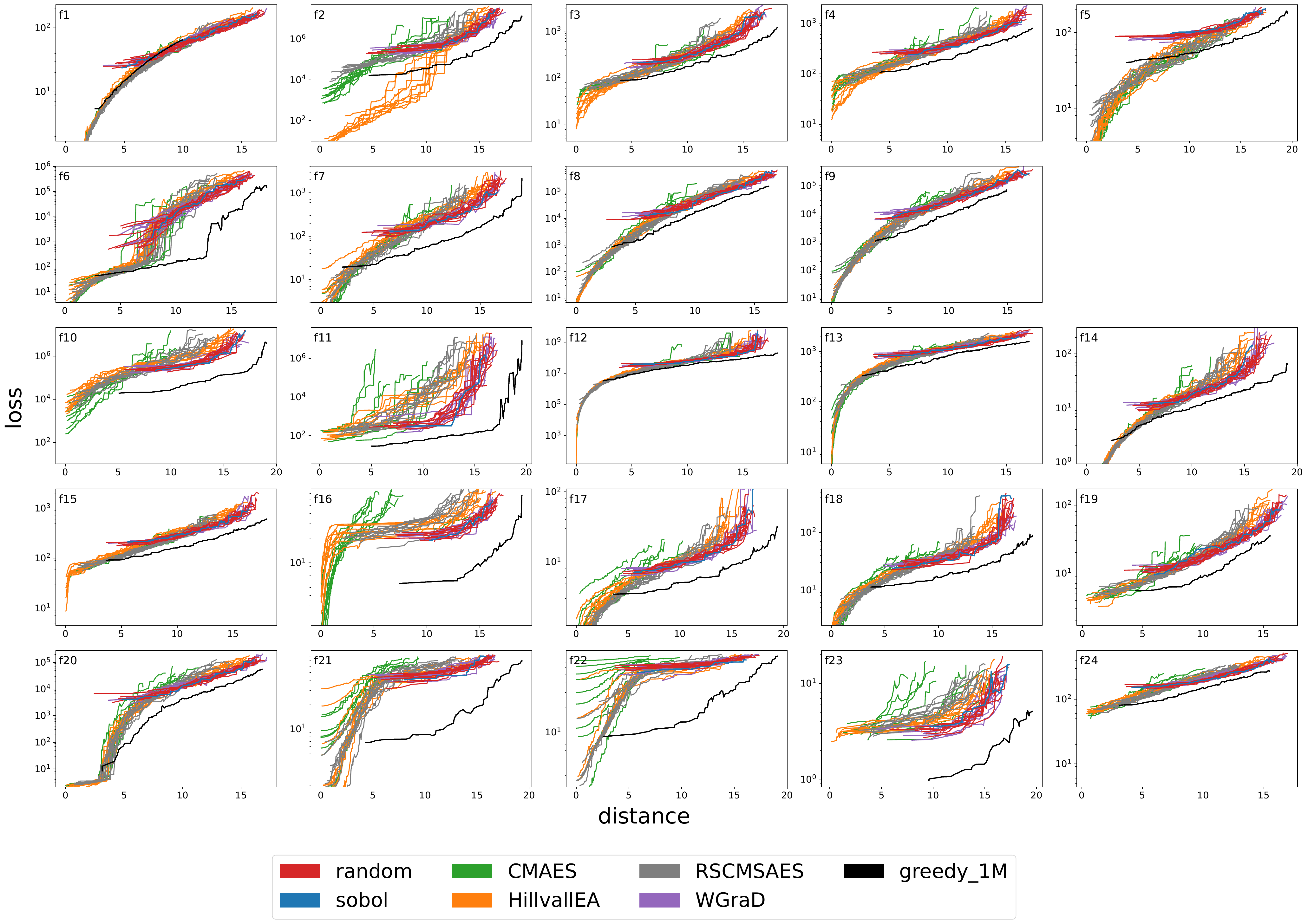}
      \caption{$D=10$. Same as Figure \ref{d10k5log}, but for initial portfolio size $T = 1\,000$.}
    \label{d10k51000log}
\end{figure}

Compared to $D=2$, differences among algorithms become more pronounced, with MMO algorithms often outperforming the one-shot strategies across all functions for at least half of the considered enforced distance range. Random sampling frequently catches up later on, as evident in Figure~\ref{d10k5log}.

Among the MMO sampling strategies, WGraD performs the least effectively. Even for the smallest distances, it locates closely positioned solutions in the search space, but of poor quality. Consequently, it proves to be an unfavorable option for the larger portfolio size $T=10\,000$. On the other hand, HillvallEA, RS-CMSA, and CMA-ES exhibit strong performance, producing point histories with many points in the strongest basins of attraction compared to the one-shot sampling strategies, which suffer from sparseness in dimension 10.
This sparseness is also underscored by the delayed onset of the curves obtained from the random sampling strategy on the x-direction, irrespective of the initial portfolio size: $T=10\,000$ in Figure~\ref{d10k5log}, $T=1\,000$ in Figure~\ref{d10k51000log}, or 1 million of points. This is because the points comprising the batch of optimal loss are already far apart from each other from the first iteration of our greedy algorithm. This contrasts with MMO algorithms and CMA-ES, whose nature drives them to locate many more and closer points in the major basins of attraction.

The gap between the algorithms and the \textit{greedy\_1M} baseline becomes more evident than in dimension 2, especially in Figure~\ref{d10k51000log}, where the trade-off curves are generated starting from a portfolio of 1\,000 points. 
The difference between the algorithms and the black curve (\textit{greedy\_1M}) observed in Figure~\ref{d10k51000log} compared to Figure~\ref{d10k5log} reinforces the discussion in Section~\ref{sec:K5}, emphasizing that higher-dimensional spaces clearly benefit from a larger initial portfolio size.

This indicates that for smaller portfolio sizes, it is essential that the subset selection is performed over a portfolio that is carefully generated, providing a good base for identifying solutions in promising areas, despite the portfolio's reduced capability to satisfactorily cover the space.

\begin{figure}[h!]
    \center
    \includegraphics[width=\textwidth]{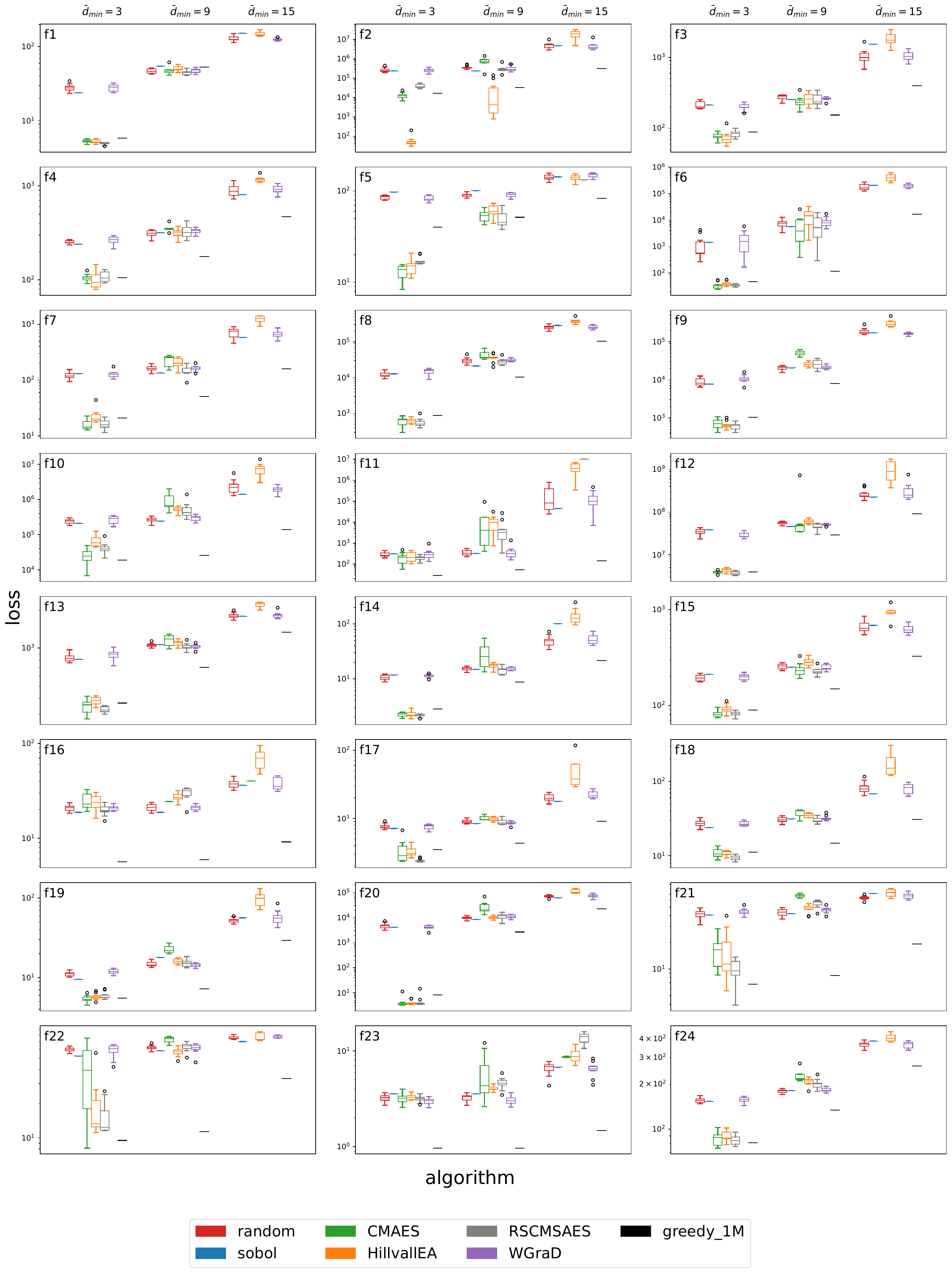}
      \caption{$D=10$. Box plots illustrating the performance of different algorithms at selected enforced distances ($\bar{d}_{\text{min}} = 3$,  $\bar{d}_{\text{min}} = 9$, and  $\bar{d}_{\text{min}} = 15$) for a batch
of $k = 5$ points. 10 independent runs of the greedy approach starting from different point portfolios of size $T = 1\,000$.}
    \label{boxplotd10k51k}
\end{figure}

To support the discussion, Figure~\ref{boxplotd10k51k} presents box plots illustrating the performance of various algorithms at three different enforced distance levels, in the setting $D=10$, $k=5$, and $T=1\ 000$. 

Additional box plots for other settings are available in the supplementary material.
The box plots in Figure~\ref{boxplotd10k51k} clearly show that, for small distances ($\bar{d}_{\text{min}} = 3$), MMO algorithms and CMA-ES outperform non-adaptive sampling strategies across all analyzed functions. However, as the distance constraint increases ($\bar{d}_{\text{min}} = 9$), random sampling and Sobol sampling begin to achieve comparable or even better performance than the other algorithms, eventually dominating at $\bar{d}_{\text{min}} = 15$. This aligns with the search nature of the algorithms used to generate the portfolios: if the focus is on identifying one or multiple optimal solutions, as with CMA-ES and MMO algorithms respectively, they will tend to underexplore less promising areas of the search space, regardless of their diversity. 
Indeed, we are unable to find points at minimum distance $\bar{d}_{\text{min}} = 15$ within the trajectories of RS-CMSA and CMA-ES. For CMA-ES, we find a valid batch only for function f23, suggesting that modifying this algorithm to explicitly account for the distance constraint could improve its applicability to our problem.
For WGraD we observe the following: when given a budget of $T=1\,000$ function evaluations, we struggle to select from its trajectory good batches for small enforced pairwise distance. However, its relative performance improves as the distance requirement is strengthened, eventually matching the quality of the batches obtained from the non-adaptive sampling strategies. This can be attributed to WGraD’s design, which uses weighted gradient and distance-based clustering to detect niches by constructing spanning trees. Results~\citep{9002742} have shown that it explores high-dimensional spaces more effectively than classical niching methods. This suggests its potential for real-world applications with high-dimensional search spaces and limited evaluation budgets.

\begin{figure}[t]
    \center
    \includegraphics[width=\textwidth]{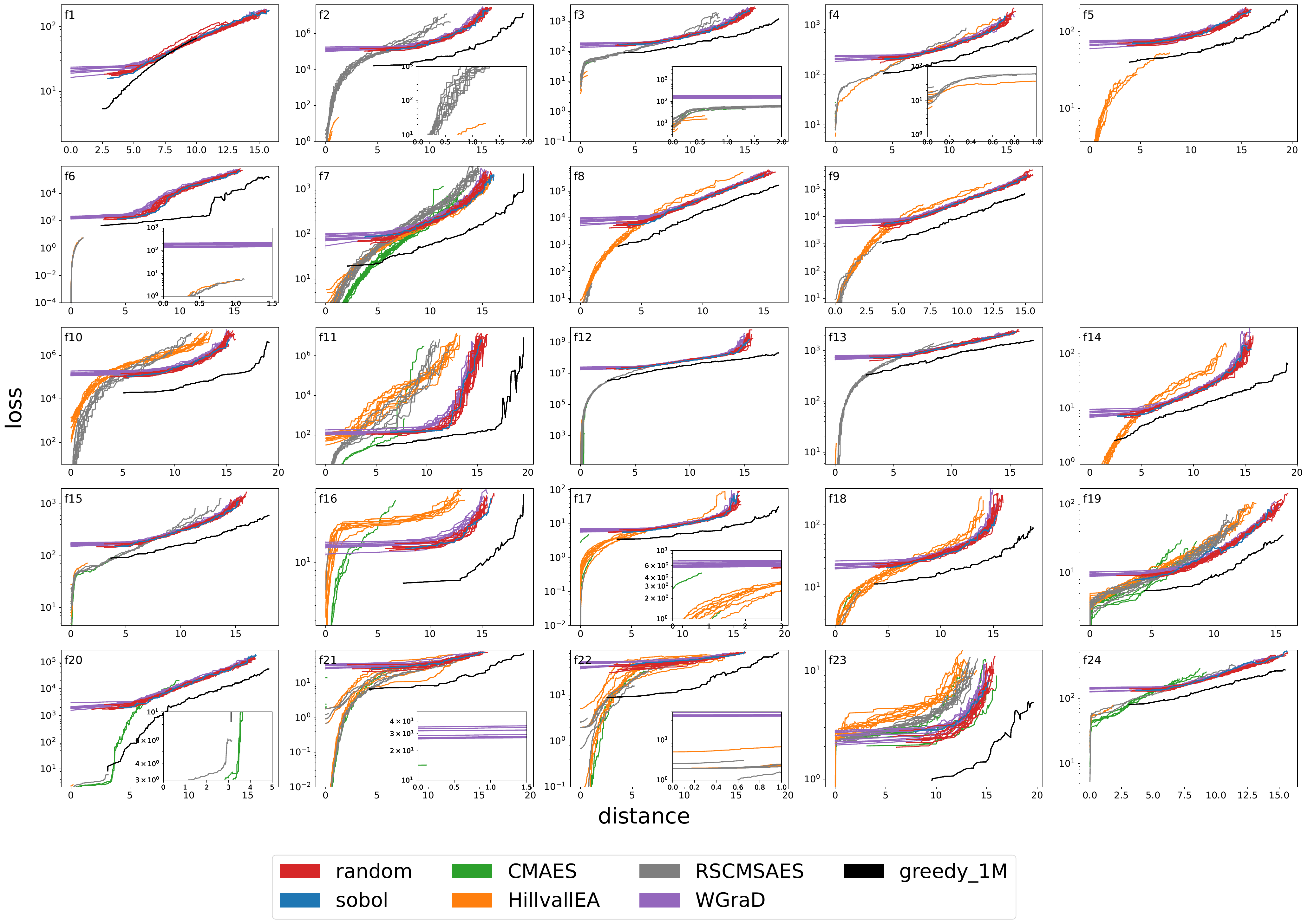}
      \caption{$D=10$. Trade-off between loss and minimum distance for an optimal batch of $k = 10$ points. 10 independent runs of the greedy approach starting from different point portfolios of size $T = 10\,000$. Insets in the bottom right corner highlight functions where some runs of the HillvallEA, RS-CMSA, and CMA-ES algorithms make little progress in both axes.}
    \label{d10k10log}
\end{figure}

For $D = 10$, given the larger volume of the search space, it is also worth considering in more detail the results for batch size $k = 10$. 
In Figure~\ref{d10k10log}, we see that some runs of the HillvallEA, RS-CMSA, and CMA-ES algorithms make very little progress in both distance and loss (e.g., f2-f4, f6, f17, f20-f22). The insets in the bottom right corner of the figure highlight these details. This occurs because, with $T=10\,000$, these algorithms find many solutions that are close to the true optimum and to each other. The greedy algorithm updates the batch of solutions with only small increases in the average pairwise distance. Additionally, the points that are substituted within the batch are very close to the ones that were already there due to the presence of many near-optimal solutions in the algorithms' history. Consequently, it takes many evaluations of the greedy algorithm before we notice a significant worsening of the average pairwise distance in the batch and an increase in the minimum pairwise distance. This could be mitigated by requiring the greedy algorithm to increase the minimal distance between any two points in the batch by some $\varepsilon>0$ at each step or by simply increasing the number $M$ of iterations. However, given a good understanding for the general trends can be obtained from the results plotted in the figures, and for resource-conscious experimentation, we do not explore this alternative for this work but rather keep it in mind for future experiments.

\begin{figure}[h!]
    \center
    \includegraphics[width=\textwidth]{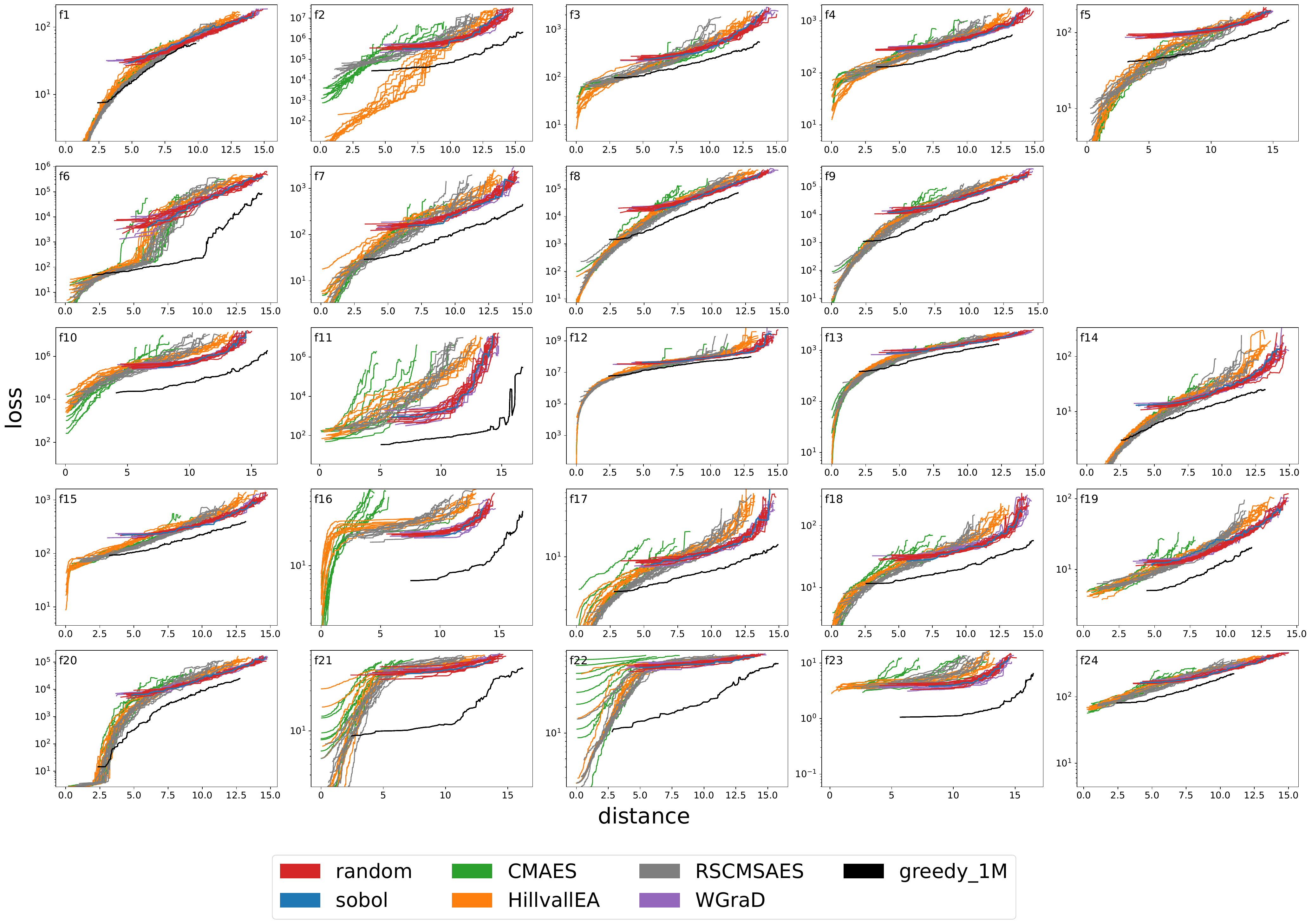}
      \caption{$D=10$. Same as Figure \ref{d10k10log}, but for initial portfolio size $T = 1 000$.}
    \label{d10k101000log}
\end{figure}

\section{Conclusion and Future Perspectives}
\label{sec:conclusion}

In real-world optimization problems, additional criteria beyond the primary objective, e.g., manufacturability, aesthetics, and robustness, are often extremely difficult---if not impossible---to translate into precise analytical definitions and hence to be integrated into the optimization procedure. To enable qualitative post-hoc evaluation, it can hence be of practical importance to return \textit{batches} of solutions that are diverse in terms of designs.

In this study, we explored the effectiveness of established algorithms in generating trajectories from which optimal subsets can be selected to balance average quality and diversity in the search space. 
We benchmarked multi-modal optimization algorithms, established heuristics, and non-adaptive sampling strategies by iteratively extracting high-quality, dispersed solution batches.
Our evaluation covered the 24 noiseless BBOB functions from the COCO environment, spanning dimensions from 2 to 10, and using different portfolio and batch sizes. 

By evaluating different portfolios of sizes 1\,000 and 10\,000 samples, we observed that CMA-ES tends to generate trajectories lacking solution diversity, aligning with its single-objective global optimization purpose. Consequently, its performance on our task heavily depends on the specific run. In contrast, MMO strategies exhibit better trajectory diversity and higher stability across different runs for a specific setting. While it is possible to extract good batches for small enforced pairwise distances, the average quality of the best batches is lower than for those obtained from the portfolio of randomly sampled points. 

These results underscores the potential for developing algorithms to effectively address our diversity-fitness challenge.

Furthermore, extracting solution sets balancing average quality and diversity revealed consistent trends across various function groups, thereby enhancing our understanding of the landscapes under consideration.

Our analysis highlights several key takeaways: (1)~ our greedy approach for extracting point batches that balance quality and input space diversity has proven to be an effective tool for analyzing and understanding black-box function landscapes; (2)~ on some multimodal landscapes, looking for optimal solutions naturally ensure distance constraints to be satisfied, while other functions require explicit enforcement; (3)~ algorithms such as CMA-ES and MMO strategies show limitations under large distance constraints---CMA-ES often lacks diversity due to its primary focus on global optimization, while MMO struggles with stronger constraints, particularly when basins of attraction are located close to one another; (4)~ introducing diversity-enforcement mechanisms into global optimizers like CMA-ES, while maintaining their ability to identify high-quality global optima, represents a promising, viable, and underexplored research direction; and (5)~ generating diverse solution batches is critical for real-world applications, where additional criteria such as manufacturability, production costs, or aesthetics are challenging to formalize analytically, emphasizing the need for algorithms that balance diversity and fitness effectively.

With our study, we aim to generate interest in the research community regarding our fundamental question: \textit{What trade-off exists between the minimum distance within batches of solutions and the average quality of their fitness?} As we gain a clearer understanding, we question whether the subset selection problem can indeed be transformed into the development of algorithms tailored for such a purpose.

\subsection*{Acknowledgments} This work was realized with the financial support of the Adelaide-CNRS Research Mobility Scheme (006016334), of the Sorbonne Center for Artificial Intelligence (SCAI) of Sorbonne University (IDEX SUPER 11-IDEX-0004), of ANR project ANR-22-ERCS-0003-01, of the CNRS INS2I project \emph{IOHprofiler}, and of the Australian Research Council (ARC) through grant FT200100536.

\bibliographystyle{apalike}
\bibliography{references}
\end{document}